\documentclass[lettersize,journal]{IEEEtran}
\usepackage{amsmath,amsfonts}
\usepackage{algorithmic}
\usepackage{algorithm}
\usepackage{array}
\usepackage[caption=false,font=normalsize,labelfont=sf,textfont=sf]{subfig}
\usepackage{textcomp}
\usepackage{stfloats}
\usepackage{url}
\usepackage{verbatim}
\usepackage{graphicx}
\usepackage{booktabs}
\usepackage{multirow}
\usepackage{booktabs}
\usepackage{tabularx}
\usepackage{multirow}
\usepackage{makecell}
\usepackage{array}
\usepackage{enumitem}
\usepackage[table]{xcolor}

\usepackage{colortbl}

\definecolor{rankone}{RGB}{255, 230, 153}   
\definecolor{ranktwo}{RGB}{217, 217, 217}   
\definecolor{rankthree}{RGB}{244, 204, 204} 

\definecolor{impactone}{RGB}{255,235,238}   
\definecolor{impacttwo}{RGB}{255,243,224}   
\definecolor{impactthree}{RGB}{255,249,196} 
\usepackage{cite}
\usepackage[colorlinks=true, linkcolor=blue, citecolor=blue, urlcolor=blue]{hyperref}

\begin{document}

\title{GPA-VGGT:Adapting VGGT to Large Scale Localization by Self-Supervised Learning with Geometry and Physics Aware Loss}


\author{IEEE Publication Technology,~\IEEEmembership{Staff,~IEEE,}}
\author{Yangfan Xu,  Lilian Zhang,  Xiaofeng He,Yugui Shen, Pengdong Wu,Wenqi Wu, Jun Mao*
}



\maketitle
\begin{abstract}
Transformer-based visual geometry models have shown strong potential for camera pose estimation and 3D scene understanding from video streams. Among them, Visual Geometry Grounded Transformer (VGGT) provides a powerful framework for joint camera and geometry modeling, but its training typically depends on large amounts of ground-truth supervision, which limits its applicability to unlabeled and unseen large-scale environments. In this paper, we present a self-supervised framework to adapt VGGT for large-scale visual localization from unlabeled driving videos. Instead of relying on conventional pairwise supervision, we extend self-supervision to sequence-wise geometric constraints that exploit richer temporal and multi-view consistency within long video streams. Specifically, multiple source frames are sampled in each sequence and geometrically projected onto different target frames to reinforce cross-view feature alignment and long-range spatial consistency. We further introduce a joint optimization objective that couples photometric consistency with geometric constraints, enabling the model to learn physically grounded camera motion and scene structure without hard labels. With this training strategy, the pretrained geometric backbone of VGGT is retained, while the camera and depth prediction heads are effectively adapted to large-scale localization scenarios under the proposed self-supervised objective. Experiments on challenging large-scale outdoor benchmarks demonstrate that the proposed method consistently improves long-range localization accuracy and trajectory stability in dynamic and adverse environments..
Our code will be released at https://github.com/X-yangfan/GPA-VGGT.

\end{abstract}

\begin{figure*}[t]
    \centering
    \includegraphics[width=\textwidth]{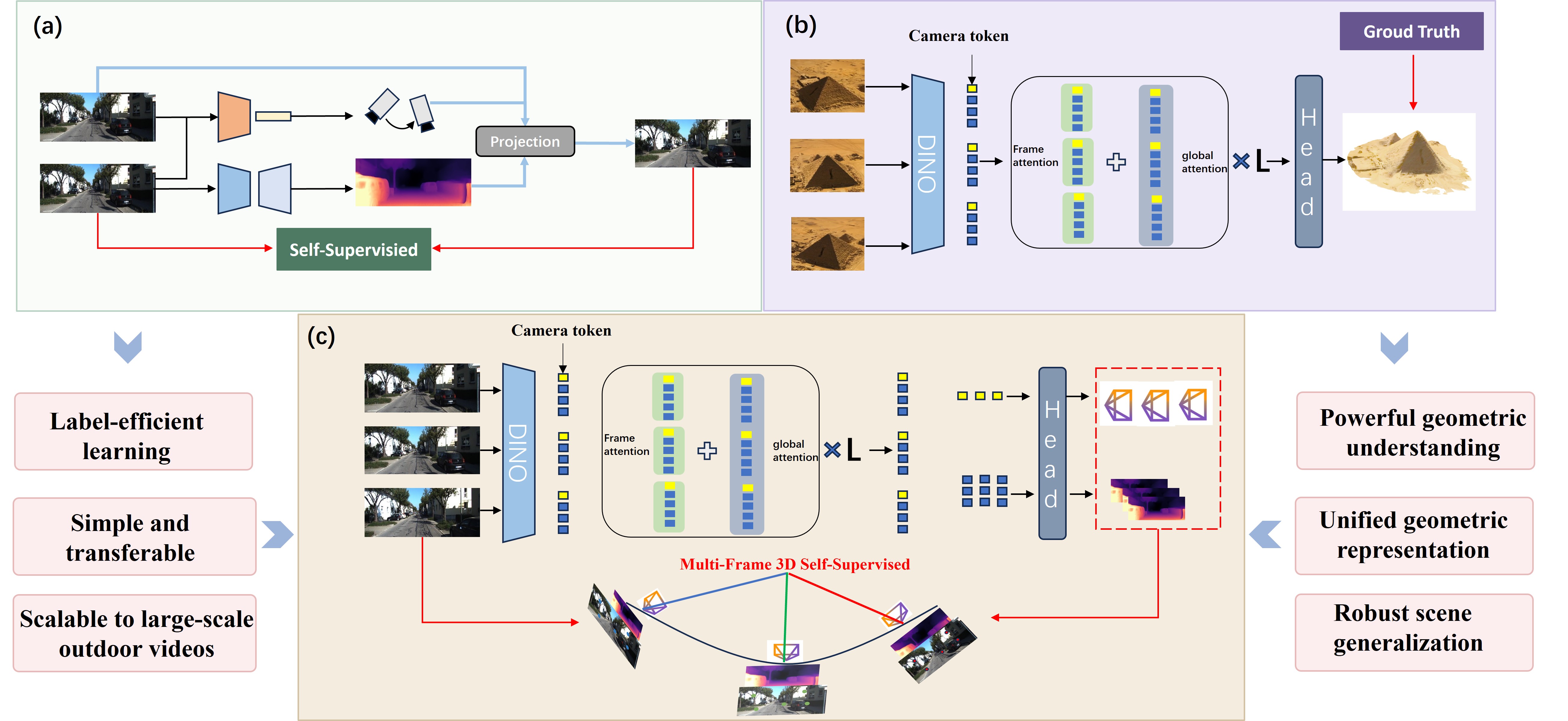}
    \caption{Comparison of traditional self-supervised learning, VGGT, and our multi-sequence 3D self-supervised framework, which induces large-scale geometric reasoning in VGGT through structured loss design without architectural changes.}
    \label{fig:image1}
\end{figure*}

\section{Introduction}

Large-scale localization and 3D reconstruction from monocular video is a fundamental capability for real-world multimedia perception, with broad applications in autonomous navigation, embodied interaction, and immersive scene understanding.

While conventional camera pose estimation and 3D reconstruction have long relied on multi-view geometry~\cite{schoenberger2016sfm,9440682}, recent advances in neural networks have enabled visual 3D tasks to be addressed in an end-to-end manner~\cite{10655144,10.1007/978-3-031-73220-1_5,Kendall_2015_ICCV,teed2021droid,tang2018banet}. In particular, attention-based architectures and large-scale pre-training have made it possible to jointly infer camera pose, depth, and correspondences from long image sequences, demonstrating strong geometric reasoning capabilities. However, these models typically rely on richly annotated training data, including accurate camera poses, depth maps, or feature correspondences, which are costly and difficult to obtain at scale in real-world environments. Consequently, although large volumes of visual data can be readily collected in practical applications, most of them remain underexploited by supervised learning, and model performance often deteriorates when deployed in domains that differ from those seen during training.

A natural way to mitigate this bottleneck is to exploit unlabeled video through self-supervision. Prior self-supervised methods attempted to alleviate this dependence on geometric annotations by learning depth and camera motion from photometric consistency. While such methods remove the need for explicit supervision, their training signals are predominantly derived from short-range frame-to-frame relationships, making it difficult to enforce geometric consistency over long video sequences. Moreover, they are inherently sensitive to dynamic objects, occlusions, and illumination changes, which further limit their robustness in real-world scenarios. These limitations suggest that the challenge lies not only in the absence of supervision, but also in the limited ability of earlier approaches to model long-range geometry. More recently, geometry foundation models, such as DUSt3R~\cite{10655144}, VGGT~\cite{wang2025vggt}, and their sequential extensions~\cite{maggio2025vggt-slam,wang2025continuous,chen2026tttr}, have demonstrated substantially stronger capabilities in global geometric reasoning. Among them, VGGT is particularly well suited for self-supervised adaptation on long monocular videos, as it jointly models camera pose, depth, and geometric correspondences within a unified transformer-based framework, providing a stronger basis for enforcing long-horizon geometric consistency. This progress creates new opportunities for exploiting long unlabeled videos through self-supervised adaptation or post-training. Nevertheless, how to effectively post-train such a model on long unlabeled videos, while maintaining robustness to dynamic objects, occlusions, and other sources of noisy supervision, remains largely underexplored.

To address this limitation, what is needed is a self-supervised adaptation strategy that can effectively exploit the long-range geometric structure in unlabeled videos while remaining robust to the noisy supervision induced by real-world dynamics. To this end, we propose a self-supervised post-training framework that adapts the Visual Geometry Grounded Transformer (VGGT) to large-scale localization using unlabeled video data. Our method combines a sequence-level geometric consistency objective within temporal windows with a physics-aware supervision mechanism that incorporates motion-consistent temporal weighting and geometry-guided source selection. Concretely, supervision is encouraged to favor source-target relations that are more consistent with plausible temporal motion and cross-view geometry, while suppressing unreliable signals caused by dynamic objects, occlusions, and abrupt appearance changes. In this way, the model enables more stable long-horizon geometric learning from unlabeled videos, leading to improved localization accuracy and generalization in large-scale environments. Figure~\ref{fig:image1} illustrates the differences between traditional self-supervised learning, the original VGGT framework, and our proposed sequence-level self-supervised adaptation strategy for long-range geometric reasoning.

Our contributions are summarized as follows:
\begin{itemize}
    \item We present a self-supervised post-training framework for adapting VGGT to large-scale localization and depth prediction from unlabeled videos, improving its performance in real-world deployment without ground-truth supervision.
    \item We propose a self-supervised long-sequence geometry learning scheme that combines sequence-level geometric consistency within temporal windows with physics-aware hard selection, improving long-horizon scale and trajectory coherence under dynamic and occluded conditions.
    \item We conduct extensive experiments on multiple large-scale datasets, demonstrating consistent improvements in global localization accuracy and generalization under challenging environmental conditions.
\end{itemize}
\section{Related Work}

\subsection{Classical Geometry-based Localization and Reconstruction}

Camera pose estimation and scene reconstruction have long been studied through classical geometric pipelines, particularly Structure-from-Motion (SfM) and Simultaneous Localization and Mapping (SLAM). These methods rely on explicit geometric modeling, feature detection and matching, bundle adjustment, and multi-view geometric optimization, and they have laid the foundation for modern visual localization and 3D reconstruction systems~\cite{hartley2003multiple,10.1145/1141911.1141964,schoenberger2016sfm,schoenberger2016sfm,Engel-et-al-pami2018}. Representative systems such as ORB-SLAM and ORB-SLAM3 further demonstrate that geometry-driven optimization remains highly effective for accurate and robust localization across monocular, stereo, RGB-D, and visual-inertial settings~\cite{murAcceptedTRO2015,9440682}.

Despite their strong performance, classical pipelines typically depend on carefully engineered modules, hand-crafted priors, and fragile intermediate correspondences. Their performance can degrade in large-scale outdoor environments with dynamic objects, significant appearance changes, repetitive structures, and severe environmental noise. These limitations have motivated the development of learning-based approaches, which aim to infer scene geometry and camera motion directly from visual observations while reducing the reliance on manually designed components and explicit correspondence estimation~\cite{Kendall_2015_ICCV,zhou2017unsupervised,teed2021droid,tang2018banet}.

\subsection{Self-supervised Depth and Ego-motion Learning from Video}

Self-supervised learning has become a mainstream paradigm for estimating scene depth and camera ego-motion from monocular video, as it removes the need for dense ground-truth supervision for depth or pose. A seminal line of work formulates the problem as view synthesis, where depth and relative camera motion are learned by minimizing photometric reconstruction errors between adjacent frames~\cite{zhou2017unsupervised,8100182}. Building on this formulation, subsequent methods improve performance by introducing stronger geometric constraints and more robust training objectives. For instance, Mahjourian \emph{et al.} enforce 3D geometric consistency by aligning temporally adjacent point clouds~\cite{mahjourian2018unsupervised}, while Monodepth2 improves depth quality and training stability through minimum reprojection loss and auto-masking strategies~\cite{monodepth2}. Later approaches such as PackNet-SfM, ManyDepth, and SC-Depth further enhance monocular self-supervised learning through improved network architectures, multi-frame reasoning, and additional geometric regularization~\cite{packnet,watson2021temporal,sc_depthv3}.

Despite the advantage of label-free training, most existing methods still rely primarily on pairwise supervision or short temporal windows~\cite{watson2021temporal,zeng2024rsa}. Consequently, geometric constraints are mainly imposed across local neighboring frames, and long-range temporal consistency over extended trajectories remains insufficiently explored~\cite{video_depth_anything,11092714}. In addition, many of these approaches are built upon CNN-based backbones, whose inherently local receptive fields limit their ability to capture global scene context and sequence-level geometric structure. These limitations often lead to scale inconsistency, accumulated drift, and reduced robustness when deployed in large-scale outdoor environments with complex motion patterns, dynamic objects, and substantial appearance variations~\cite{10.1609/aaai.v33i01.33018001}.

\subsection{Foundation Models for 3D Perception and Visual Localization}

Recent advances in foundation-model-style vision architectures have increasingly influenced 3D perception, localization, and mapping. General-purpose self-supervised models such as DINO and DINOv2 show that large-scale pretraining can produce robust and transferable visual representations across diverse downstream tasks, providing strong front-end priors for geometry-related reasoning~\cite{caron2021emerging,oquab2023dinov2}. In parallel, depth-oriented foundation models, including Depth Anything and Depth Anything V2, demonstrate that scaling data, model capacity, and pretraining objectives can substantially improve the robustness and generalization of monocular depth estimation~\cite{depth_anything_v1,depth_anything_v2,piccinelli2024unidepth,piccinelli2025unidepthv2,ke2023repurposing,ke2025marigold}. These developments suggest that large pretrained models are becoming an increasingly important backbone for downstream 3D understanding systems.

Building on this trend, recent geometry foundation models move beyond generic visual representation learning toward explicit multiview geometric reasoning. DUSt3R reformulates 3D reconstruction from uncalibrated images through dense point-map prediction, while MASt3R further strengthens geometry-aware matching and improves downstream reconstruction, pose estimation, and localization performance~\cite{10655144,10.1007/978-3-031-73220-1_5}. VGGT pushes this direction further by unifying camera estimation, depth prediction, point-map regression, and 3D tracking within a single feed-forward Transformer, highlighting the potential of general-purpose geometric inference under a unified architecture~\cite{wang2025vggt}.

At the system level, several recent works explore how such geometry foundation models can be extended to larger-scale and longer-horizon scenarios. VGGT-SLAM integrates VGGT predictions into a full SLAM pipeline for dense RGB mapping and localization, while VGGT-Long improves scalability on kilometer-scale monocular sequences through chunk-based processing, overlap alignment, and lightweight loop closure~\cite{maggio2025vggt-slam,deng2025vggtlongchunkitloop}. Related extensions also investigate multimodal geometric inputs and streaming or incremental inference to improve online and memory-bounded 3D perception over long image streams~\cite{omnivggt2025,zhuo2025streaming4dvisualgeometry,fang2026incvggt,yan2026omnistreammasteringperceptionreconstruction}. More broadly, recent long-video depth models such as Video Depth Anything further emphasize that maintaining geometric consistency over extended sequences remains a central challenge for large-scale visual perception~\cite{depth_anything_v1}.

Despite this progress, an important gap remains for self-supervised visual localization in realistic outdoor settings. Existing self-supervised localization and geometry-learning methods are still largely dominated by local frame-pair supervision or short temporal windows, while recent foundation-model-based approaches mainly emphasize representational strength, multimodal extension, or system scalability rather than label-free adaptation to large-scale unlabeled localization data. In contrast, our work focuses on adapting a high-capacity geometry foundation model to self-supervised outdoor localization by introducing sequence-level geometric consistency and a physics-aware hard selection strategy, enabling more stable learning under dynamic objects, occlusions, and abrupt appearance changes.

\begin{figure*}
    \centering
    \includegraphics[width=2\columnwidth]{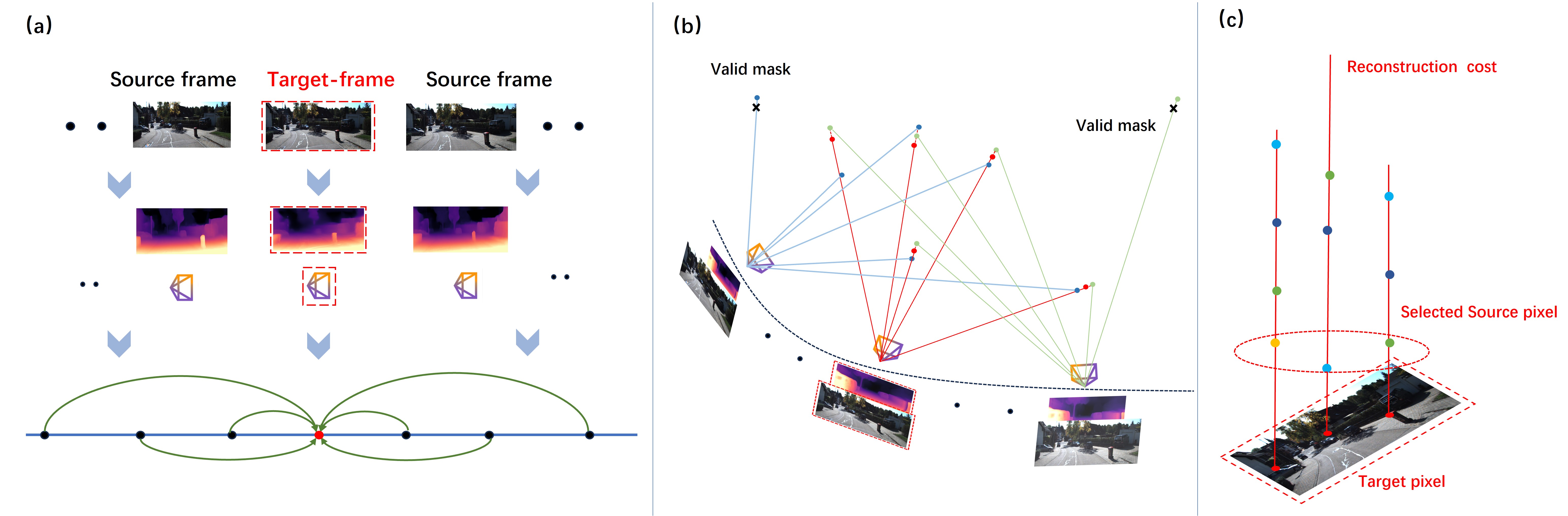}
    \caption{(a) Illustration of target and source frame assignment within a temporal window in the framework. 
    (b) Multi-frame projection from source frames to a keyframe with validity mask construction. 
    (c) Per-pixel hard source selection, where the source view yielding the minimum photometric--geometric cost is selected for supervision.}
    \label{fig:image2}
\end{figure*}

\section{Method}
\label{sec:method}

\subsection{Network Architecture and Parameterization}

Given a temporal window
\begin{equation}
\mathcal{W}=\{I_0, I_1, \dots, I_{S-1}\},
\end{equation}
our framework jointly estimates scene depth, camera intrinsics, and frame poses in an end-to-end manner. The input sequence is first encoded into shared multi-scale spatio-temporal features, which are then fed into a camera branch and a depth branch.

The camera branch predicts the scene-level intrinsic matrix \(\mathbf{K}\) together with the pose of each frame relative to the first frame of the sliding window:
\begin{equation}
\mathbf{T}_{0\rightarrow i}
=
[\mathbf{R}_{0\rightarrow i}\,|\,\mathbf{t}_{0\rightarrow i}]
\in SE(3),
\end{equation}
where frame \(0\) serves as the local reference coordinate system of the current window. Since intrinsics are fixed within the same scene but may vary across scenes, \(\mathbf{K}\) is estimated at the scene level rather than independently per frame. To ensure stable optimization, focal lengths and principal points are constrained to physically plausible ranges.

The depth branch predicts a normalized inverse-depth representation together with a pixel-wise confidence map. Metric depth is then recovered using a standard inverse-depth parameterization widely adopted in self-supervised monocular depth estimation \cite{monodepth2}.

\subsection{Bidirectional Temporal Formulation and View Synthesis}

We treat each temporal window as a sparse view graph of the same rigid 3D scene and introduce a \emph{Bidirectional Uniform Geometric Coverage Strategy} to impose denser cross-view supervision over the full window. For each window, a subset of frames
\begin{equation}
\mathcal{T}=\{t_1,\dots,t_K\}
\end{equation}
is sampled as geometric anchors. In the forward direction, anchor frames are treated as targets and the remaining frames as sources; in the reverse direction, non-anchor frames are treated as targets and anchor frames as sources. This bidirectional construction increases geometric coverage and ensures that each frame is constrained by multiple viewpoints.

Although the network predicts poses relative to the first frame of the window, view synthesis is performed between arbitrary target-source pairs. For any pair \((t,s)\), the relative transformation from frame \(t\) to frame \(s\) is recovered by pose composition:
\begin{equation}
\mathbf{T}_{t\rightarrow s}
=
\mathbf{T}_{0\rightarrow s}\mathbf{T}_{0\rightarrow t}^{-1}.
\end{equation}
Target-view reconstruction is then performed by standard differentiable warping based on predicted depth, camera intrinsics, and relative pose \cite{zhou2017unsupervised}.

\subsection{Physical Consistency and Geometry-Guided Multi-Source Selection}

Our supervision combines photometric consistency and cross-view geometric consistency. The photometric term follows the standard SSIM+\(L_1\) formulation commonly used in self-supervised view synthesis, while cross-view geometry is enforced by comparing transformed depth and reprojected source depth following established geometric-consistency formulations.

To couple appearance consistency with geometric reliability, we use the geometric term to modulate the photometric reconstruction loss:
\begin{equation}
\tilde{\mathcal{L}}_{\text{photo}}^{(s)}
=
\mathcal{L}_{\text{photo}}^{(s)}
\cdot
\left(1-\mathcal{L}_{\text{geo}}^{(s)}\right).
\end{equation}
This reduces the influence of pixels with inconsistent 3D structure.
To account for dynamic content and the increasing uncertainty of long-range temporal matching, we further introduce an exponential temporal decay factor:
\begin{equation}
w_{t,s}=\exp\big(-\alpha(|t-s|-1)\big).
\end{equation}
The source-view reconstruction cost is defined as
\begin{equation}
\mathcal{C}_s(\mathbf{p})
=
w_{t,s}\tilde{\mathcal{L}}_{\text{photo}}^{(s)}(\mathbf{p})
+
\lambda_{\text{geo}}w_{t,s}\mathcal{L}_{\text{geo}}^{(s)}(\mathbf{p}).
\end{equation}
Instead of averaging over all source views, we adopt a per-pixel minimum-cost selection strategy:
\begin{equation}
\mathcal{L}_{\text{raw\_final}}(\mathbf{p})
=
\min_{s\in \mathcal{W}\setminus\{t\}}
\mathcal{C}_s(\mathbf{p}),
\end{equation}
which allows the model to automatically suppress occluded or geometrically inconsistent sources.

\subsection{Outlier Rejection and Confidence Modeling}

To suppress errors caused by dynamic objects and ambiguous texture-less regions, we employ a standard auto-masking mechanism that discards pixels better explained by identity reconstruction than by motion-based warping \cite{zhou2017unsupervised}. 

In addition, the network predicts a confidence map to scale the final reconstruction loss:
\begin{equation}
\mathcal{L}_{\text{final}}(\mathbf{p})
=
\mathbf{C}_t(\mathbf{p})\mathcal{L}_{\text{raw\_final}}(\mathbf{p}).
\end{equation}
To avoid the trivial solution \(\mathbf{C}_t\rightarrow 0\), we regularize the confidence prediction with a negative log penalty:
\begin{equation}
\mathcal{L}_{\text{conf}}=\mathbb{E}\big[-\log \mathbf{C}_t\big].
\end{equation}

\subsection{Overall Objective}

The total training objective combines the masked reconstruction loss, a standard edge-aware smoothness prior on inverse depth, the confidence regularization term, and auxiliary pose-related regularization:
\begin{equation}
\mathcal{L}_{\text{total}}
=
\frac{1}{|\mathcal{V}|}
\sum_{\mathbf{p}\in\mathcal{V}}
\mathcal{L}_{\text{final}}(\mathbf{p})
+
\lambda_s\mathcal{L}_{\text{smooth}}
+
\lambda_c\mathcal{L}_{\text{conf}}
+
\mathcal{L}_{\text{pose\_aux}}.
\end{equation}
Here \(\mathcal{V}\) denotes the valid pixel set after masking. The auxiliary pose terms improve motion consistency and geometric stability through weak intrinsic regularization, temporal smoothness, and \(SE(3)\) cycle consistency.

\subsection{Inference}

\begin{figure}
    \centering
    \includegraphics[width=1\columnwidth]{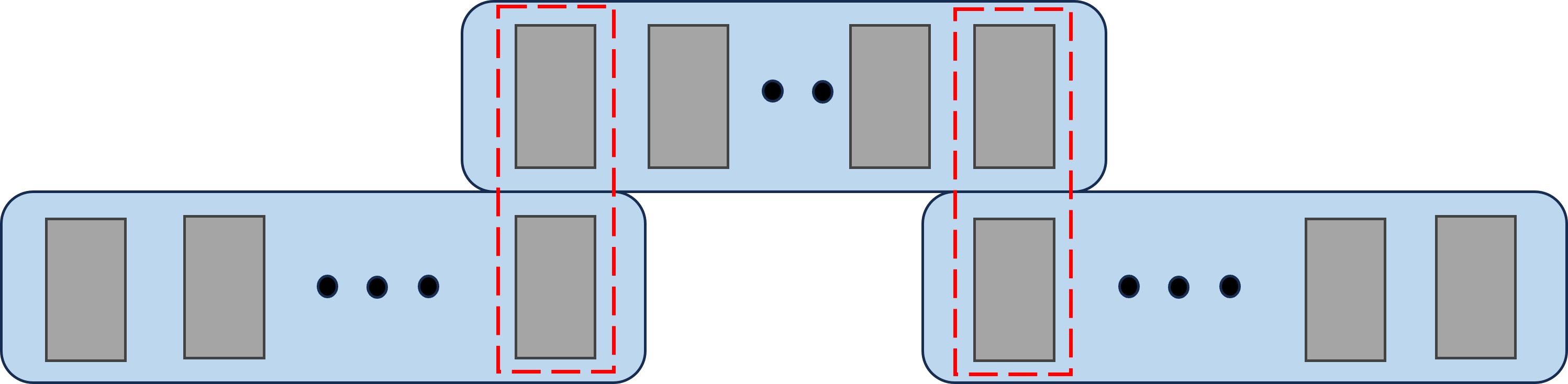}
    \caption{Chained inference with overlapping windows, where shared frames enable forward propagation of pose predictions to form long-range trajectories.}
    \label{fig:image3}
\end{figure}

During inference, long videos are processed using overlapping sliding windows. For each window, the network predicts metric depth together with frame poses relative to the first frame of that window. Relative motion between arbitrary frames is recovered by pose composition, and shared frames between consecutive windows enable chained propagation for long-range trajectory generation, as illustrated in Fig.~\ref{fig:image3}. Global trajectory recovery then follows standard pose chaining across adjacent frames.

\section{Experiments}
\label{sec:experiments}

\begin{figure*}
    \centering
    \includegraphics[width=2\columnwidth]{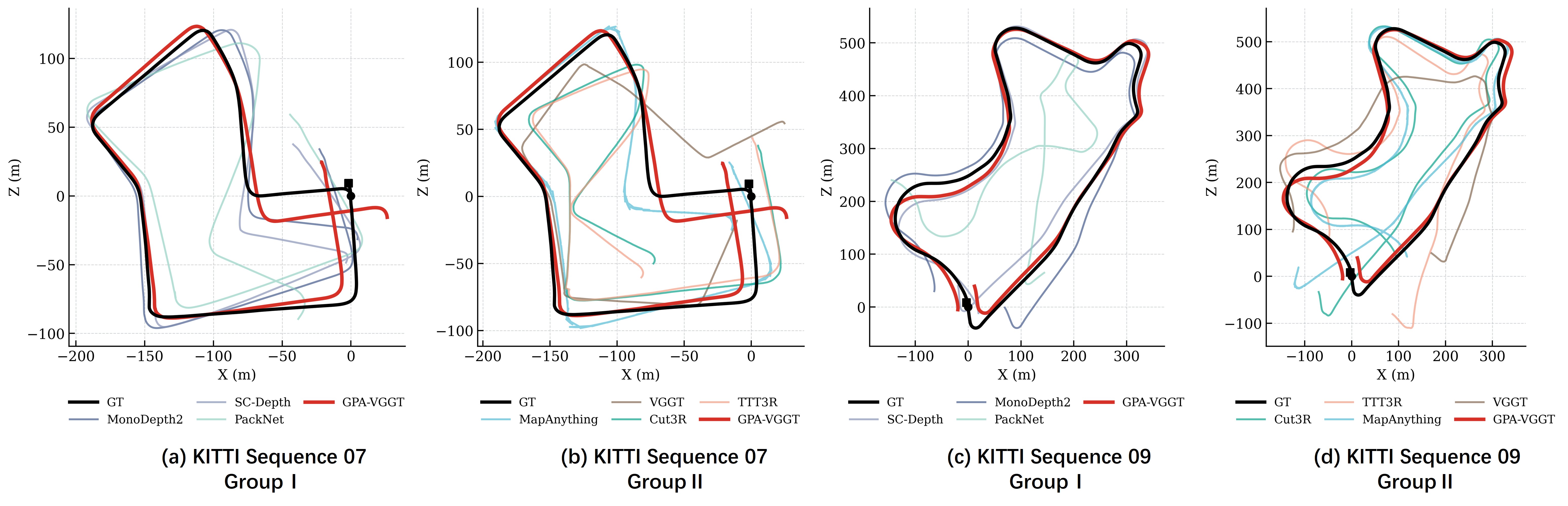}
    \caption{Trajectory comparison results on KITTI Odometry: (a) Sequence 07; (b) Sequence 09.}
    \label{fig:image4}
\end{figure*}

\subsection{Experimental Setup}

\begin{table}[t]
\centering
\caption{Adaptation study on KITTI Odometry Sequences 07 and 09. 
``VGGT'' denotes the original pretrained model without retraining. 
``Full'' loads the full pretrained checkpoint; ``w/o Attn'' removes pretrained attention-related modules; ``Only DINO'' initializes from the DINO backbone only; ``KITTI-trained'' denotes a VGGT variant conventionally trained on KITTI; and ``w/o Heads'' removes pretrained prediction heads, corresponding to our GPA-VGGT setting. 
Metrics are Absolute Trajectory Error (ATE, m) and Relative Pose Error (RPE, m); lower is better. Best results are shown in bold.}
\label{tab:vggt_adaptation_kitti}
\resizebox{\linewidth}{!}{
\begin{tabular}{lcccc}
\toprule
\multirow{2}{*}{Method} & \multicolumn{2}{c}{Sequence 07} & \multicolumn{2}{c}{Sequence 09} \\
\cmidrule(lr){2-3} \cmidrule(lr){4-5}
& ATE$\downarrow$ & RPE$\downarrow$ & ATE$\downarrow$ & RPE$\downarrow$ \\
\midrule

\rowcolor{gray!15}
\multicolumn{5}{l}{\textbf{Without retraining}} \\
VGGT & 30.507 & 0.152 & 98.568 & 0.363 \\

\rowcolor{gray!15}
\multicolumn{5}{l}{\textbf{With retraining}} \\
VGGT + Full             & 13.726 & 0.095 & 21.433  & \textbf{0.147} \\
VGGT + w/o Attn         & 75.682 & 0.586 & 172.252 & 1.038 \\
VGGT + Only DINO        & 79.348 & 0.578 & 174.060 & 1.060 \\
VGGT + KITTI-trained    & 14.863 & 0.095 & 22.952  & 0.156 \\
\rowcolor{red!8}
VGGT + w/o Heads (Ours) & \textbf{11.931} & \textbf{0.092} & \textbf{19.694} & 0.148 \\

\bottomrule
\end{tabular}
}
\end{table}

\begin{figure}
    \centering
    \includegraphics[width=1\columnwidth]{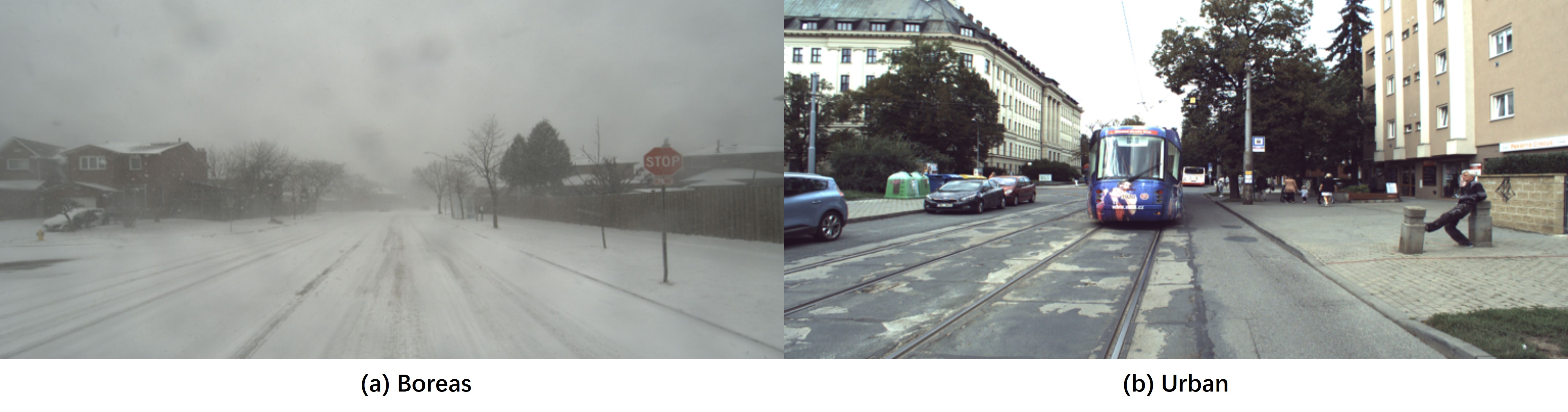}
    \caption{(a) Snowy scene from the Boreas dataset. (b) Dynamic object scene from the Urban dataset.}
    \label{fig:image7}
\end{figure}

\paragraph{Datasets.}
We evaluate our method on three outdoor benchmarks: \textbf{KITTI Odometry}~\cite{Geiger2013IJRR}, \textbf{Brno Urban Dataset}~\cite{9197277}, and \textbf{Boreas}~\cite{doi:10.1177/02783649231160195}. 
KITTI (Sequences 07 and 09) is used to assess long-range trajectory estimation. 
Urban contains highly dynamic scenes with dense traffic and pedestrians, while Boreas presents severe seasonal and weather variations, including snow, illumination changes, and low-texture regions. 
Together, these datasets evaluate robustness to long-range drift, dynamic interference, and appearance degradation. Representative examples from Urban and Boreas are shown in Fig.~\ref{fig:image7}.

\paragraph{Baselines.}
We compare with two groups of methods: \emph{monocular self-supervised approaches}, including PackNet-SfM~\cite{packnet}, SC-DepthV3~\cite{sc_depthv3}, and Monodepth2~\cite{monodepth2}; and \emph{geometry foundation models}, including VGGT~\cite{wang2025vggt}, Streaming VGGT~\cite{zhuo2025streaming4dvisualgeometry}, InfiniteVGGT~\cite{yuan2026infinitevggt}, MapAnything~\cite{keetha2025mapanythinguniversalfeedforwardmetric}, VGGSfM~\cite{10655838}, DUSt3R~\cite{10655144}, Cut3R~\cite{wang2025continuous}, and TTT3R~\cite{chen2026tttr}. 
All methods are evaluated without backend optimization.

\paragraph{Evaluation Metrics.}
We report \emph{Absolute Trajectory Error} (ATE) and \emph{Relative Pose Error} (RPE), both in meters. 
ATE measures global trajectory consistency, while RPE evaluates local motion accuracy.

\paragraph{Implementation Details.}
Input RGB images are resized to $518\times518$ and normalized before being fed into the network. 
Dataset intrinsics are used for geometric projection. 
We apply standard photometric augmentations, including brightness and contrast jittering and horizontal flipping, consistently across frames within each sampled window.

\paragraph{Training and Inference Protocol.}
We initialize the model from pretrained VGGT weights and re-initialize the prediction heads before GPA training. 
During inference, we generate long-range trajectories using a chained overlapping sliding-window protocol (Fig.~\ref{fig:image3}), where overlapping frames are used to align adjacent windows and maintain temporal continuity. 
We further adopt the acceleration strategy of FastVGGT~\cite{shen2025fastvggt}, achieving up to 18.4 FPS.

For controlled comparison, we evaluate geometry foundation baselines under the same overlapping-window inference protocol with window size $S=10$ whenever applicable. Our goal is not to reproduce the individually optimized system-level configuration of each baseline, but to compare front-end geometric prediction quality under a unified long-range chaining interface. Since directly applying many short-context or frame-based models to long monocular sequences often leads to unstable or severely degraded trajectories, the shared sliding-window protocol serves as a practical normalization step for comparison. By standardizing the temporal context, overlap pattern, and trajectory composition rule across methods, this protocol reduces the influence of backend optimization, handcrafted post-processing, and dataset-specific engineering choices. Under this setting, the evaluation primarily focuses on front-end geometric prediction quality and sequence-level chaining stability. Detailed training configurations, including the warmup setting and the multi-keyframe training protocol, are provided in the supplementary material.

\begin{figure*}[t]
    \centering
    \includegraphics[width=2\columnwidth]{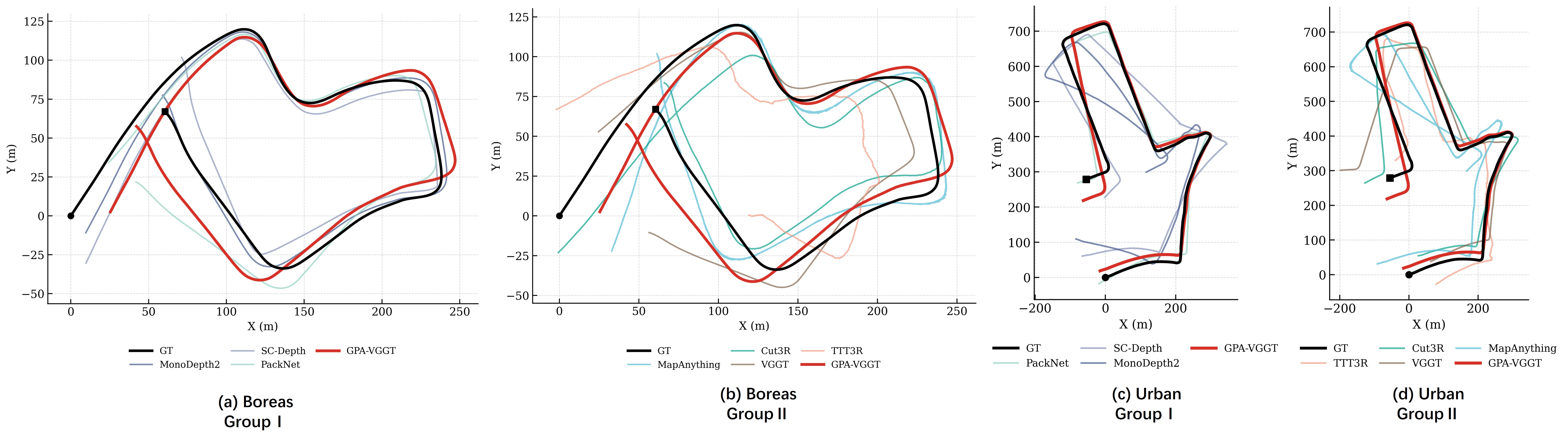}
    \caption{Qualitative trajectory comparison on the Boreas and Urban datasets. While competing methods suffer from severe scale drift and trajectory fragmentation under snowy and adverse-weather conditions, GPA-VGGT maintains a globally consistent trajectory that remains closely aligned with the ground truth.}
    \label{fig:image6}
\end{figure*}

\subsection{Comparative Studies}
\subsubsection{Analysis of Initialization and Adaptation Strategy.}
To further understand where the improvement comes from, we perform an adaptation study on KITTI, as summarized in Table~\ref{tab:vggt_adaptation_kitti}. 
Directly applying the pretrained VGGT without retraining leads to large trajectory errors on both sequences, indicating that the original model does not transfer well to long-range monocular localization under our inference setting. 
Retraining from the full pretrained checkpoint already brings substantial gains, confirming the value of VGGT's pretrained geometric prior. 
However, removing the pretrained attention-related modules or initializing from the DINO backbone only causes severe performance degradation, showing that the geometry-specific pretrained modules are crucial for stable sequence-level reasoning. 
Among all variants, re-initializing the prediction heads while loading the pretrained VGGT weights achieves the best overall performance. 
This suggests that retaining the pretrained geometric backbone while allowing the prediction heads to adapt to the GPA objective provides a more effective initialization for long-horizon localization.

\subsubsection{Comparison on KITTI Odometry}

We first evaluate the proposed method on the KITTI Odometry benchmark. As shown in Table~\ref{tab:table1}, our method achieves the best ATE on both Sequence 07 and Sequence 09. The qualitative comparisons in Fig.~\ref{fig:image4} further show that our predicted trajectories align more closely with the ground truth over long distances.

Compared with monocular self-supervised baselines, our method significantly reduces long-range drift. Although existing methods can produce reasonable short-range motion estimates, their errors accumulate progressively along the trajectory, especially on Sequence 09. Recent geometry foundation models exhibit a similar limitation: despite competitive local pose estimation, their performance degrades under chained sliding-window inference, resulting in noticeably worse global consistency. These results confirm that improving local geometric prediction alone is insufficient for large-scale monocular localization, and that explicit sequence-level consistency is critical for stable long-horizon trajectory estimation.
\begin{table}[t]
\centering
\footnotesize
\caption{Quantitative camera pose estimation comparison on KITTI Odometry Sequences 07 and 09. Metrics are Absolute Trajectory Error (ATE) and Relative Pose Error (RPE) in meters. Lower is better. 
\colorbox{rankone}{1st} \, 
\colorbox{ranktwo}{2nd} \, 
\colorbox{rankthree}{3rd}.}
\label{tab:table1}
\resizebox{\columnwidth}{!}{
\begin{tabular}{l|cc|cc}
\hline
\multirow{2}{*}{Method} & \multicolumn{2}{c|}{Sequence 07} & \multicolumn{2}{c}{Sequence 09} \\
\cline{2-5}
                        & ATE $\downarrow$ & RPE $\downarrow$ & ATE $\downarrow$ & RPE $\downarrow$ \\
\hline\hline
\multicolumn{5}{l}{\textit{Monocular Self-Supervised Methods (Group I)}} \\
\hline
PackNet        & 40.673 & 1.364 & 152.530 & 1.944 \\
SC-DepthV3     & 27.224 & 1.339 & \cellcolor{ranktwo}\textbf{23.174}  & 2.168 \\
Monodepth2     & \cellcolor{rankthree}\textbf{17.654} & \cellcolor{rankone}\textbf{0.057} & \cellcolor{rankthree}\textbf{38.916} & \cellcolor{rankone}\textbf{0.075} \\
\hline
\multicolumn{5}{l}{\textit{Supervised Geometry Foundation Models (Group II)}} \\
\hline
VGGT           & 30.507 & 0.152 & 98.568  & 0.363 \\
Streaming VGGT & 52.887 & 2.072 & 182.924 & 1.598 \\
InfiniteVGGT   & 87.900 & 2.964 & 204.373 & 15.436 \\
MapAnything    & \cellcolor{ranktwo}\textbf{14.580} & 1.423 & 63.994  & 2.474 \\
VGGSfM         & 79.756 & 0.658 & 82.696  & 0.222 \\
DUSt3R         & -      & -     & 47.458  & 3.504 \\
Cut3R          & 37.703 & \cellcolor{rankthree}\textbf{0.112} & 55.878  & 0.179 \\
TTT3R          & 40.806 & 0.114 & 62.283  & \cellcolor{ranktwo}\textbf{0.123} \\
\hline
\textbf{GPA-VGGT (Ours)} & \cellcolor{rankone}\textbf{11.931} & \cellcolor{ranktwo}\textbf{0.092} & \cellcolor{rankone}\textbf{19.694} & \cellcolor{rankthree}\textbf{0.148} \\
\hline
\end{tabular}
}
\end{table}

\subsubsection{Comparison in Challenging Real-World Environments}

We further evaluate the proposed method on the Urban and Boreas datasets to assess robustness under dynamic objects and severe appearance degradation. Quantitative results are reported in Table~\ref{tab:comparison_brno_boreas_full}, and qualitative trajectory comparisons are shown in Fig.~\ref{fig:image6}. Our method achieves the best ATE on both datasets, reaching 24.343\,m on Urban and 14.801\,m on Boreas.

The performance gain over conventional self-supervised methods is particularly clear in these challenging settings. On Urban, dense traffic and pedestrians violate the static-scene assumption and introduce substantial noise into photometric supervision. On Boreas, snow-covered roads and weak textures make image matching much less reliable. In both cases, competing self-supervised methods suffer from stronger drift or reduced stability, while our method remains substantially more robust.

Recent geometry foundation models also show degraded performance in these environments. Although they benefit from large-scale pre-training, their predictions become less stable when applied to long-range sequential inference under strong domain shifts, such as dynamic urban scenes or adverse weather. As shown in Fig.~\ref{fig:image6}, several baseline methods produce fragmented or severely drifting trajectories, whereas our method maintains much better global coherence. These results indicate that target-sequence adaptation is critical for robust localization when the test environment differs significantly from the training distribution.

\begin{figure*}
    \centering
    \includegraphics[width=2\columnwidth]{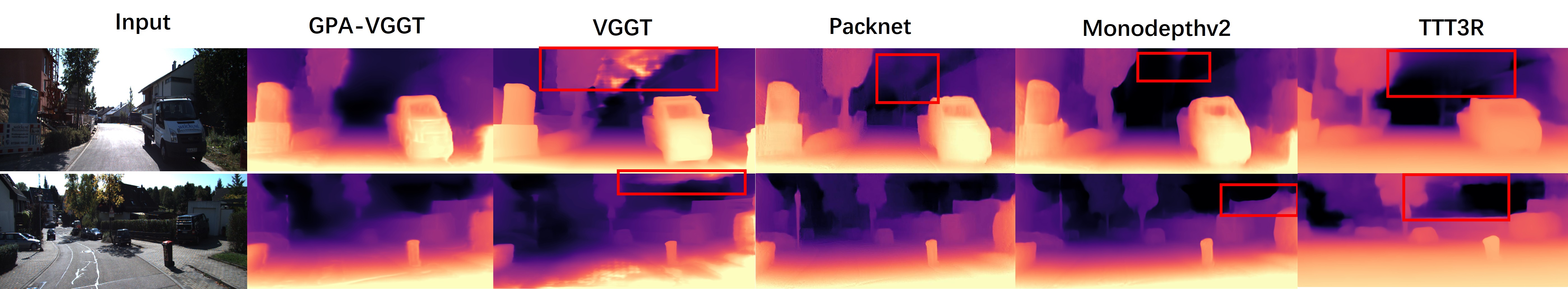}
    \caption{Comparison of depth predictions from different methods. Our method produces temporally more consistent and geometrically more coherent depth predictions across frames. The regions highlighted by the red boxes indicate typical failure-prone areas, where competing methods often suffer from artifacts such as noisy sky regions, blurry boundaries, or locally inconsistent geometry.}
    \label{fig:image5}
\end{figure*}
\begin{figure}
    \centering
    \includegraphics[width=1\columnwidth]{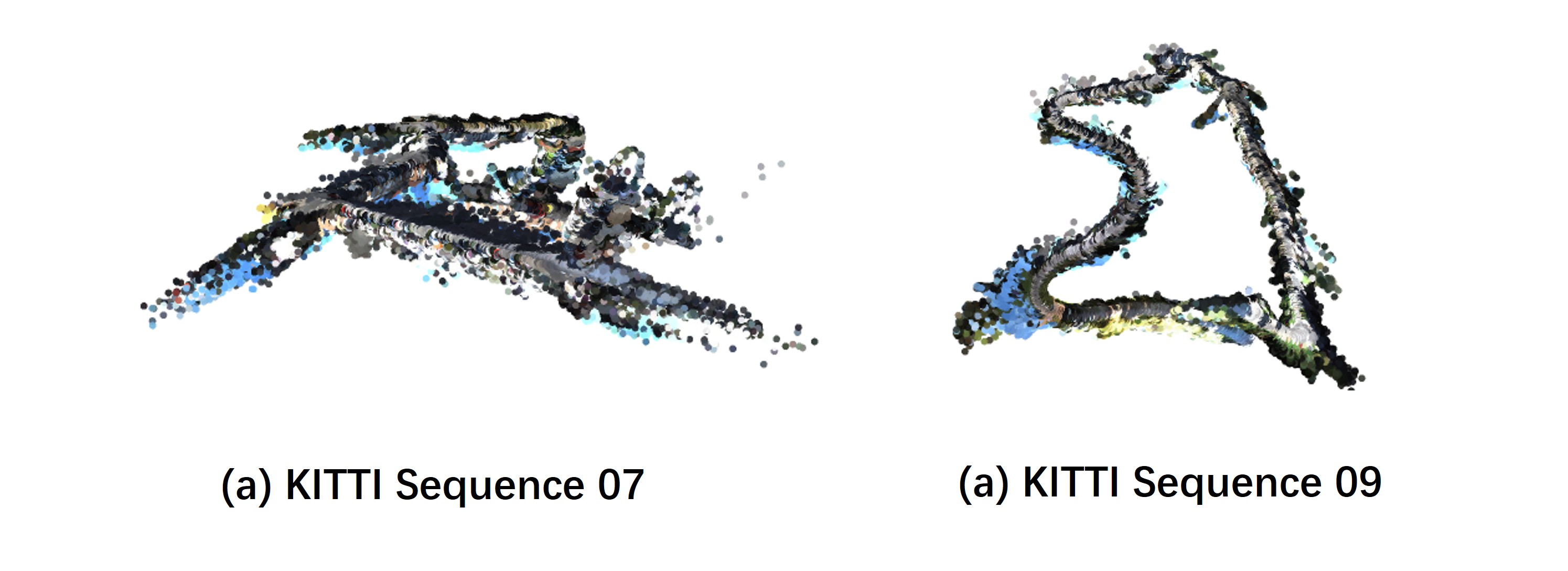}
    \caption{(a) Reconstruction results on KITTI sequence 07. (b) Reconstruction results on KITTI sequence 09.}
    \label{fig:image8}
\end{figure}

\begin{table}[t]
\centering
\footnotesize
\caption{Quantitative camera pose estimation comparison on the Urban (Dynamic) and Boreas (Snow) datasets. Metrics are Absolute Trajectory Error (ATE) and Relative Pose Error (RPE) in meters. Lower is better. 
\colorbox{rankone}{1st} \,
\colorbox{ranktwo}{2nd} \,
\colorbox{rankthree}{3rd}.}
\label{tab:comparison_brno_boreas_full}
\resizebox{\columnwidth}{!}{
\begin{tabular}{l|cc|cc}
\hline
Method & \multicolumn{2}{c|}{Urban (Dynamic)} & \multicolumn{2}{c}{Boreas (Snow)} \\ 
\cline{2-5}
       & ATE $\downarrow$ & RPE $\downarrow$ & ATE $\downarrow$ & RPE $\downarrow$ \\ 
\hline\hline
\multicolumn{5}{l}{\textit{Monocular Self-Supervised Methods (Group I)}} \\
\hline
PackNet        & \cellcolor{rankthree}\textbf{30.834}  & 1.224   & 19.869  & 0.701   \\ 
SC-DepthV3     & 59.049  & 1.232   & \cellcolor{ranktwo}\textbf{17.841}  & 0.758   \\ 
Monodepth2     & 96.275  & 1.358   & \cellcolor{rankthree}\textbf{18.831}  & 0.697   \\ 
\hline
\multicolumn{5}{l}{\textit{Supervised Geometry Foundation Models (Group II)}} \\
\hline
VGGT           & 70.792  & \cellcolor{rankone}\textbf{1.125}   & 30.258  & \cellcolor{rankthree}\textbf{0.654}   \\ 
Streaming VGGT & 119.403 & 4.653   & 56.109  & 9.984   \\ 
InfiniteVGGT   & 206.689 & 14.319  & 76.303  & 3.282   \\ 
MapAnything    & 89.426  & 2.994   & 18.623  & 1.778   \\ 
VGGSfM         & 120.140 & 1.195   & 76.652  & \cellcolor{rankone}\textbf{0.553}   \\ 
DUSt3R         & \cellcolor{ranktwo}\textbf{25.064}  & 2.792   & 40.763  & 4.239   \\ 
Cut3R          & 63.405  & \cellcolor{rankthree}\textbf{1.181}   & 25.600  & 0.747   \\ 
TTT3R          & 54.411  & 1.190   & 40.241  & \cellcolor{ranktwo}\textbf{0.630}   \\ 
\hline
\textbf{GPA-VGGT (Ours)} & \cellcolor{rankone}\textbf{24.343} & \cellcolor{ranktwo}\textbf{1.151} & \cellcolor{rankone}\textbf{14.801} & 0.736 \\
\hline
\end{tabular}
}
\end{table}

\subsection{Ablation Study}
\label{sec:ablation}

We conduct ablation studies on KITTI Sequences 07 and 09 to evaluate the main design choices of GPA-VGGT. The results are reported in Table~\ref{tab:ablation_07_09}. Among all variants, the three most influential ablations are \emph{w/o geometric weighting}, \emph{Keyframe: only 2 nearest sources}, and \emph{Adjacent-pair supervision only}, which directly validate the importance of our geometry-aware weighting, multi-source keyframe supervision, and non-local temporal supervision design.

\paragraph{Geometry-guided weighting.}
Removing geometric weighting causes the largest degradation, especially on Sequence 09, where the ATE increases dramatically from 21.433\,m to 62.777\,m. This result shows that directly applying photometric supervision without geometry-aware filtering introduces severely unreliable gradients in occluded or inconsistent regions. Therefore, geometry-guided weighting is a core component for stabilizing self-supervised optimization in long-range localization.

\paragraph{Multi-source keyframe supervision.}
Restricting supervision to only the two nearest source frames also leads to a major performance drop, increasing the ATE on Sequence 09 from 21.433\,m to 43.372\,m. This indicates that sparse local keyframe constraints are insufficient for stable long-horizon trajectory estimation. In contrast, using multiple source frames provides richer and more complementary geometric constraints, which is essential for reducing accumulated drift.

\paragraph{Beyond adjacent-pair supervision.}
Limiting supervision to adjacent frame pairs degrades the ATE to 15.724\,m on Sequence 07 and 41.815\,m on Sequence 09. This result suggests that pairwise local supervision alone cannot adequately constrain the full temporal window, even when local motion estimation remains relatively accurate. Instead, broader temporal supervision is necessary to improve sequence-level consistency over long trajectories.

Overall, these ablations consistently show that the gain of GPA-VGGT does not come from a single loss term, but from the combination of three key ideas: reliable geometry-aware supervision, multi-source keyframe constraints, and temporally extended sequence-level learning. Together, these components substantially improve global trajectory consistency, especially on the longer and more challenging Sequence 09.

\begin{table}[t]
\centering
\footnotesize
\setlength{\tabcolsep}{5pt}
\renewcommand{\arraystretch}{1.05}
\caption{Ablation study of GPA-VGGT on KITTI Odometry Sequences 07 and 09. Metrics are Absolute Trajectory Error (ATE) and Relative Pose Error (RPE) in meters. Lower is better. \colorbox{impactone}{1st} \, \colorbox{impacttwo}{2nd} \, \colorbox{impactthree}{3rd} denote the three most influential ablation settings selected in our analysis.}
\label{tab:ablation_07_09}
\resizebox{\columnwidth}{!}{
\begin{tabular}{l|cc|cc}
\hline
\multirow{2}{*}{Method} & \multicolumn{2}{c|}{Sequence 07} & \multicolumn{2}{c}{Sequence 09} \\
\cline{2-5}
 & ATE $\downarrow$ & RPE $\downarrow$ & ATE $\downarrow$ & RPE $\downarrow$ \\
\hline\hline

\multicolumn{5}{l}{\textit{Base}} \\
\hline
GPA-VGGT (Ours) & \textbf{11.931} & 0.092 & \textbf{19.694} & 0.148 \\
\hline

\multicolumn{5}{l}{\textit{Basic Ablations}} \\
\hline
\rowcolor{impactthree}
Adjacent-pair supervision only & 15.724 & 0.090 & 41.815 & 0.152 \\
Mean source aggregation & 12.784 & 0.090 & 24.798 & 0.144 \\
w/o depth consistency & 12.725 & 0.095 & 33.881 & 0.150 \\
\hline

\multicolumn{5}{l}{\textit{Reviewer-requested Ablations}} \\
\hline
\rowcolor{impacttwo}
Keyframe: only 2 nearest sources & 12.296 & 0.088 & 43.372 & 0.153 \\
Softmin aggregation ($\tau=0.02$) & 17.657 & 0.105 & 28.291 & 0.160 \\
w/o auto-mask & 13.032 & 0.092 & 40.150 & 0.150 \\
\rowcolor{impactone}
w/o geometric weighting & 15.223 & 0.097 & 62.777 & 0.183 \\
Forward-only supervision ($\lambda_{rev}=0$) & 14.298 & 0.100 & 42.306 & 0.159 \\
w/o smoothness regularization & 11.968 & 0.090 & 23.213 & 0.150 \\
L1 photometric loss only (w/o SSIM) & 12.249 & 0.091 & 30.312 & 0.145 \\
\hline
\end{tabular}
}
\end{table}

\subsection{Qualitative Analysis}

We qualitatively compare our method with representative baselines on both standard and challenging benchmarks. On KITTI, the trajectory comparisons in Fig.~\ref{fig:image4} show that GPA-VGGT follows the ground-truth path more closely on Sequences 07 and 09, with reduced long-range error accumulation and fewer abrupt deviations under chained propagation. These observations are consistent with the ATE improvements reported in Table~\ref{tab:table1}, suggesting that the proposed sequence-level supervision improves global trajectory stability.

The advantage becomes more evident on challenging datasets. In Boreas, snowy weather, illumination variation, and large low-texture regions make visual matching and photometric reasoning less reliable, while in Urban, frequent dynamic objects such as vehicles and pedestrians often break temporal consistency. As shown in Figs.~\ref{fig:image6} and~\ref{fig:image7}, several baselines suffer from fragmented trajectories or severe drift in these cases, whereas GPA-VGGT maintains smoother and more coherent long-horizon predictions.

Figure~\ref{fig:image5} presents qualitative depth comparisons. Rather than emphasizing frame-wise sharpness alone, we focus on temporal geometric consistency across adjacent frames. Compared with self-supervised baselines without depth priors, our method produces cleaner and more stable depth predictions, especially in distant or weakly textured regions such as the sky, road boundaries, and thin structures. We further visualize reconstruction results on KITTI in Fig.~\ref{fig:image8}, where our method better preserves the overall scene layout and cross-frame geometric consistency.

Overall, these qualitative results support our main claim: for long-range monocular localization, enforcing sequence-level geometric consistency is more important than optimizing frame-wise predictions alone.

\section{Conclusion}

In this work, we present a geometry-aware self-supervised training framework for large-scale monocular localization based on VGGT. By enforcing consistency in sequential geometric predictions and multi-view reprojection relationships, the proposed objective improves long-range trajectory accuracy and geometric consistency without modifying the model architecture.

Experiments on KITTI, Urban, and Boreas show that our method consistently improves long-range localization, particularly in dynamic and adverse environments. These results suggest that, for large-scale 3D perception, effective geometric supervision is as important as model capacity.

Our current framework still has limitations. Long-range inference relies on chained propagation over overlapping windows, which can accumulate errors when local alignment becomes unreliable. Performance may also degrade in extremely challenging scenes with heavy occlusion, severe dynamics, or weak texture.

Overall, this work shows that large-scale geometric reasoning and monocular localization can be substantially improved through training objective design. We hope this perspective will support future research on more stable long-horizon inference and more scalable self-supervised 3D scene understanding.
\bibliographystyle{IEEEtran}
\bibliography{ref}

\end{document}